\documentclass[hidelinks]{scrartcl}
\usepackage{vhistory}
\usepackage{hyperref}
\usepackage{float}
\usepackage{amsmath}
\usepackage{breakcites}
\restylefloat{figure}
\usepackage{graphicx}
\usepackage[font=scriptsize,labelfont=bf]{caption}
\hypersetup{pdfencoding=auto}

\usepackage[textsize=scriptsize]{todonotes}

\begin{document}
	\setlength\intextsep{3pt}
	
	\newcommand{\smallheading}[1]{\par{\emph{\textbf{#1}}}}
	\title{Cognitive Development of the Web}
	\author{Viktoras Veitas (\texttt{vveitas@gmail.com}) \\
	David Weinbaum (Weaver) (\texttt{space9weaver@gmail.com})\\
	The Global Brain Institute, VUB}
	\date{August 12, 2014}
	
\maketitle
\pagestyle{plain}

\section{The Web as a sociotechnological system}
The sociotechnological system is a system constituted of human individuals and their artifacts: technological artifacts, institutions, conceptual and representational systems, worldviews, knowledge systems, culture and the whole biosphere as an evolutionary niche. In our view the sociotechnological system as a super-organism is shaped and determined both by the characteristics of the agents involved and the characteristics emergent in their interactions at multiple scales. Our approach to sociotechnological dynamics will maintain a balance between perspectives: the individual and the collective. Accordingly, we analyze dynamics of the Web as a sociotechnological system made of people, computers and digital artifacts (Web pages, databases, search engines, etc.).  

Making sense of the sociotechnological system while being part of it, is also a constant interplay between pragmatic and value based approaches. The first is focusing on the actualities of the system while the second highlights the observer's projections. In our attempt to model sociotechnological dynamics and envision its future, we take special care to make explicit our values as part of the analysis. In sociotechnological systems with a high degree of reflexivity (coupling between the perception of the system and the system's behavior), highlighting values is of critical importance\footnote{A quite radical approach is that the world is entirely constructed from specific discourses between people\cite[p. 28]{willke_smart_2007}\cite[p. 77]{heyes_selection_2001}}.

In this essay, we choose to see the future evolution of the web as facilitating a basic value, that is, continuous open-ended intelligence expansion. By that we mean that we see intelligence expansion as the determinant of the 'greater good' and 'well being' of both of individuals and collectives at all scales. Our working definition of intelligence here is \textit{the progressive process of sense-making of self, other, environment and universe}. Intelligence expansion, therefore, means \textit{an increasing ability of sense-making}.

The two most significant factors in the dynamics of the sociotechnological system is increasing complexity and accelerating change (see \cite{helbing_globally_2013, wefs_global_agenda_council_on_complex_systems_perspectives_2013} and \footnote{C. Thoeret's talk at summer school of Web Science and the Mind, Université du Québec à Montréal, July 17, 2014}). These are both consequences of technological progress. Moreover, intelligence expansion seems to lag behind changes humans introduce into the environment hence there is lack of foresight and difficulty to envision a sustainable future. As a result the overall system becomes more fragile. Pragmatically, we think that the challenge of increasing complexity can be met by intelligence expansion.

\section{Intelligence and intelligence expansion}
\subsection{Sense-making}
The concept of sense-making denotes a rich research field active since around 1970s where insights from philosophy, sociology, cognitive, information and computer science, organizational studies and more are brought together. A psychologically oriented definition of sense-making is: \textit{sensemaking is a motivated, continuous effort to understand connections (which can be among people, places, and events) in order to anticipate their trajectories and act effectively in relation to them} \cite[p. 3]{klein_1._2006}. Beside human sciences, the theory of sense-making is also used in researching human-computer interaction, decision support systems and artificial intelligence  \cite[p. 5]{klein_2._2006}.

Various authors list different aspects of sense-making. It is not our goal to provide a comprehensive list here. The most important aspects in our context are listed below and are loosely based on \cite{sensemaking_2014}:
	\begin{itemize}
	\item \textbf{Identity and identification}. A prior notion of an entity 'which is making sense' seems to be needed, but in our framework it is not the case: the identity of cognitive agents is created during the process. This extension informed by Simondon's theory of individuation\cite{simondon_genesis_1992} is central in our thesis that merges the process of emergence of boundaries that co-define individuals and their environment and the sense-making process by these formed individuals into a single process. In this essay we have chosen to define cognitive agents before sense-making only for didactic reasons.
	\item \textbf{Retrospection}. For effective sense-making, both spacial and temporal patterns (also called 'structure') should be found in the sensory data (see subsection \ref{sec:generic_cognitive_agent}). We use the terms \textit{the environment}, \textit{milieu} (environment with history/memory) and \textit{encompassing culture} interchangeably in different contexts, but they denote the same general concept of the structured 'external reality' of an agent.
	\item \textbf{Enaction}. The primary component of sense-making is action: an agent acts upon the environment, catches the 'reflection' or response of environment and updates a representation of it (not unlike a sonar). According to \cite{clark_whatever_2012} perception is an action where the agent produces a stream of expectations and only corrects its own model according to incoming information. 
	\item \textbf{Ongoing}. Sense-making is an iterative process of modeling in terms of information compression when individuals shape and react to the environments they face \cite{sensemaking_2014};
	\item \textbf{Reflexive}. Sense-making is a two-ways interaction between the individual and its environment across the boundary being created during the same process: any examination, modeling and action of agents 'bends' the environment and affects the perception and further decisions by those same agents. Furthermore, in the context of a sociotechnological system, the environment of an agent are other cognitive agents, therefore sense-making activity is always participatory: intelligence is always a product of both 'individual' and 'collective' (for the related discussion about collective individuation in social networks, see \cite{yuk_hui_collective_2013}).
	\end{itemize}
	
\subsection{Individual-environment milieu}
We maintain that the most important factor enabling to understand a sociotechnological system are interactions among intelligent agents (see section \ref{sec:framework_of_scalable_cognition}). Yet, it follows straightforwardly from our definition of intelligence as sense-making, that agents get individuated while interacting. Therefore it is important to describe how we treat a cognitive agent (see subsection \ref{sec:generic_cognitive_agent}).

In our framework an individuated agent is a kind of sensory machine which makes sense of whatever sensory information comes through its sensors. Structuring the environment in its classical understanding (i.e. taking all the data available and building an approximate model out of it) is just one option. Other options for making sense of the environment can be:
\begin{itemize}
	\item Filtering information as complete information may be less structured than part of it;
	\item Forgetting in case of mining for temporal patterns;
	\item Movement in the environment, therefore changing the immediate surrounding culture;
	\item Changing the environment itself. 	
\end{itemize}

While we talk separately about the individual and its environment (milieu), "no individual would be able to exist without a milieu, that is its complement, arising simultaneously from the operation of individuation: for this reason the individual should be seen as but a partial result of the operation bringing it forth"\cite[p. 27]{combes_gilbert_2013}. It is of fundamental importance to the framework of individuation that the formation of boundary between individual and milieu is something that is formed during the process of individuation. The implication of this principle is that the unit of analysis is neither the individual nor the environment, but the individual-milieu dyad \cite[p. 3]{simondon_genesis_1992}.

A useful framework which helps to understand the formation of individual-milieu dyad is John Holland's signal/boundary model\cite{holland_signals_2012}. Holland maintains that "Ecosystems, governments, biological cells, markets, and complex adaptive systems in general are characterized by intricate hierarchical arrangements of boundaries and signals" and that these systems can be understood only by studying the origin and co-evolution of signal / boundary hierarchies\cite[p. 12]{holland_signals_2012}. In the context of ecosystems, each new species or adaptation offers new possibilities of interaction (signals) among species (boundaries) in the ecosystem. An example of the result of such co-evolution is the fine-tuned Madagascar orchid / sphinx moth dyad \cite[p. 29]{holland_signals_2012} or "relationship of propagation between certain flowers and honeybees"\cite[p. 8]{yuk_hui_collective_2013}.

We argue that the emphasis on sense-making as an individuation process that brings forth cognitive agents in contrast to analyzing static properties of already individuated cognitive agents is more informative about the nature and prospects of intelligence expansion on the scales of human individuals, social systems, artificial intelligences, the internet and the future Web.

\subsection{Worldviews}
A worldview in the context of sense-making is a gestalt perception, both individual and collective, in relation to self,others, society and the cosmos at large. Every worldview is essentially subjective (in the sense that one cannot know what it is like to have a particular worldview without actually embracing it), or intersubjective, depending on the choices of the agents and communities sharing them. When applying the concept to the sociotechnological system, a worldview is a more or less integrated system of cognitive and behavioral patterns governing the interactions of the agent. In other words, a worldview is an integral system of sense-making characterizing an intelligent agent. In our discussion of future scenarios for the sociotechnological system, the intelligence expansion of an agent at any level is synonymous with the evolution of the worldview characterizing that agent. Hui and Halpin in their article "Collective Individuation: The Future of the Social Web" refer to the concept of a worldview by using German word \textit{Weldbild}\cite{yuk_hui_collective_2013}.

In the following section we develop in more detail our scalable framework of individuating cognitive agents.

\section{A framework for scalable cognition}
\subsection{The generic cognitive agent}\label{sec:generic_cognitive_agent}
A cognitive agent is an agent characterized by displaying cognitive activity. Cognitive activity in the broadest sense may be defined as a non-trivial derivation of actions in response to events in the agent’s environment \cite{franklin_ontology_2008, franklin_ida:_1998}. Non-trivial here means that the derivation of actions is influenced by the environment, by the situation of the agent and follows a goal or a fitness criteria. Cognitive activity also includes adaptation/learning in regard to future derivations of actions based on the success or failure of previous actions. For example: given a predefined goal or a set of fitness criteria, if the current action was a successful response of the agent to the current event, the association between the event and the selected action is reinforced, otherwise the association is weakened.

Generally, an event is any difference in the environment that affects the situation of at least one agent. An action is any effect an agent may produce in its environment. Actions therefore produce events in the environment. An agent is identified by the events that affect it, by the events it is capable to produce and by the manner the latter are associated with the first. Of course the manner of association encodes an implicit (and sometimes explicit) semantic structure; though it may include random or probabilistic elements, it is never entirely random. In other words, an agent must have a structure.

Similarly, the environment of the agent must have a structure, otherwise there can be no meaningful way for the agent to associate its actions to events since an environment without structure will necessarily respond randomly to the agent’s actions. The realization of cognitive activity must therefore assume a structural coupling \cite{maturana_organization_1975} between the agent and the environment: differences in the environment induce differences in the agent and vice-versa. Cognitive activity can be understood therefore as the manner by which an agent is affecting and being affected by the environment. Such activity has consequences on the dynamic structures of both agent and environment.

From the standpoint of an agent, the environment is the medium of events (i.e. differences) that affect it and are produced by it. When we think about a population of interacting agents, the actions of agents produce events that affect other agents and induce them to produce further actions. Agents share an environment to the extent that they interact through events, i.e. differences they exert on each other, but for each agent, all other agents are basically just environment\footnote{In contrast to this general view, when relating to the formation of coalitions of agents, as we will see in the following, it is the specific interactions among specific agents that are instrumental to their organization into coalitions.}.

Agents, in general, are not necessarily cognitive. A cognitive agent is distinguished in that it derives actions in response to events in a non-trivial manner. Clearly, some kind of information processing must be involved for any activity to be considered as cognitive. But information processing is not sufficient. If an agent’s response to events involves no selection and no dependence on a state (i.e. no memory) than it can be considered trivial. Cognitive activity therefore necessarily involves, in addition to information processing, an encoding of a situation or a state unique to the agent at a given moment and some selection mechanism that depends also on this unique state. In other words, if all the information determining an agent’s immediate action can be solely derived from the immediate event it is affected by, than, no cognitive activity is involved on the side of the agent. This conclusion is consistent with the understanding of structural coupling because it means that the structure of the agent carries no relevant information affecting its response to events, i.e. it is unchanging. If the structure of the agent is constant, it is neither affected nor affecting the structure of the environment, i.e. no structural coupling.  For cognitive agents, in contrast, there is extra information which is necessary such as the agent’s state, goals, knowledge, tendencies, etc.

Following this rationale, the humble thermostat that features in so many examples in the cybernetic literature is indeed a rudimentary cognitive agent because the knowledge of its current state is necessary to correctly predict its response to an incoming event (change in temperature). A simple electronic calculator which is much more complex than the thermostat in term of information processing is not a cognitive agent because its input sequence alone (buttons pushed) determines its actions (displaying alpha numeric characters on its display).

The question remains however whether the presence of information processing, a unique inner state and structural coupling are sufficient conditions to designate cognition. On this question there is no clear consensus. In  \cite{adams_why_2010} additional conditions are suggested having to do with the production of semantic content unique to the agent. Of a particular interest here is the following condition:  a cognitive process must involve semantic content that arise in the cognitive process and is meaningful for the agent. As we’ll see in the following, selection for relevance and the formation of challenges for the agent thereof can be considered to produce such semantic content.

A simple yet a very general working definition of cognition can be given now: cognition is the iterative coordinated processes of:
\begin{enumerate}
	\item  Selecting from the incoming stream of events which events are relevant and which are not. Relevance need not necessarily be a binary value. Events can be prioritized with varying levels of relevance according to the selection mechanism involved. The mechanism responsible for selection for relevance will referred from here on as the attention mechanism. Attention as an elementary cognitive function is exactly the singling out of relevance.
	\item  Given the current (most) relevant event, selecting from the available options of response what is the most appropriate action to execute next. An action may produce an event, change the agent’s state or do nothing. The mechanism responsible for selection for effective action will be referred from here on as the intention mechanism. Intention as an elementary cognitive function is exactly the singling out of action.
\end{enumerate}	
According to this working definition cognition is basically a selective process. Importantly, cognition as selection is not necessarily deterministic as the current selection may involve probabilistic elements in the selective processes, or, dependency on a (possibly indefinite) number of previous events not all of which can be known (i.e. the current selection depends on events older than the beginning of observation).  Any one of the selective elementary processes mentioned here can be redundant, but not both of them. If selection for relevance is redundant (i.e. no selection), either every event is relevant so the agent is maximally sensitive, or, no event is relevant so the agent is indifferent. For example: an electron in an electromagnetic field is maximally sensitive, every difference in the field is relevant to its behavior. It does not select to which differences in the electric field to respond. If selection for action is redundant (no selection), either the relevant event solely determines the action so the agent is maximally instinctive, or, no action is selected so the agent is inactive. In the example of the electron, the event of difference in the electric field already determines the electron’s response. When an agent is both maximally sensitive and maximally instinctive, like the electron in the electric field, it is not a cognitive agent anymore.

Implicit in this definition is that selection is made according to some set of criteria and possibly according to an internal state that may encode, goals,  drives,  accumulated experience of past interactions, predictive models and theories about the environment and more. These implicit elements constitute together what may be called the context of the cognitive process. The information content of events is raised to a status of semantic content as it is related (by selection) to the agent’s context. For example: an agent sharing information with her peers on a social network is not merely copying information from one location to another (information processing). The activity of sharing is related to context sensitive values such as gaining prestige and trust, inviting exchange, asking for advice etc. The so called "automatic" operation of sharing involves selection of what to share and with whom. These selections indeed give rise to semantic content that mark the activity as cognitive. In the absence of context there is no cognition. Relevance, the mark of the agent’s intelligent interaction with its environment is context sensitive. It is the agent’s dynamic situation which guides its cognitive activity. Consequently the agent’s actions affect its situation by closing a cybernetic loop through the environment.

An important aspect of our working definition of cognition is that in information theoretic terms a selection is an operation that reduces the information contained in the agent’s state. In other words, it is an irreversible operation. Once part of the information (the information content of irrelevant events for example) is lost, it is impossible to reconstruct the agent’s state prior to selection using only the information present after the selection. More in specific, the selections the agent makes in response to events are constraining the variety \cite{ashby_requisite_1958} of the effects of these events on certain parameters of its state (say its distance from a goal state).

Reducing information (variety) is how different agents gain different ‘perspectives’ in relation to the same events. Each agent reduces the incoming information in a manner that is unique to its state and goals while selecting.  Selection for relevance and selection for action, therefore, render cognition an irreversible informational process. Context dependent selections clearly seem to be the mechanism by which unique semantic content arises for the agent. The criterion of structural coupling ensures that the context itself changes in response to such selection this is how a cognitive agent ‘learns from experience’. Clearly, this poses a criterion that differentiates cognitive agents from general agents: cognitive activity is characterized by reduction of information and the development of context sensitive perspectives (semantic content). Following Ashby’s discussion of requisite variety, a population of cognitive agents with diverse perspectives may achieve with coordinated activity higher fitness than is achievable by any single agent because they can together reduce more variety from the effects of the environment\footnote{This is a good starting point to establish why coalitions of diverse cognitive agents can achieve more than any of the participating agents. It also alludes on how the measure the advantage of coalitions.}.

From yet another perspective, cognition and most clearly the selection for relevance are symmetry breaking operations: prior to selection an input set of events may seem equivalent but post selection events can be regrouped according to relevance into smaller equivalence sets. The initial symmetry is broken \cite{heylighen_growth_1999}. This notion of symmetry breaking is important in order to understand how the selections of the agent impart structure on the environment (see next section).

\subsection{Scalable cognition}\label{sec:framework_of_scalable_cognition}
The working definition of cognition suggested in section above is inspired by Bernard Baars’ global workspace theory of consciousness \cite{baars_cognitive_1993, baars_global_2005} and Stan Franklin’s application of the theory in his work on the ontology of cognition \cite{franklin_foundational_2006,franklin_ontology_2008} and artificial minds \cite{franklin_artificial_1995}. Yet, our definition aims to highlight different aspects of the cognitive process in order to prepare the ground for a scalable framework for cognition. In this, it also draws from the concept of adaptive agent and cognition as an adaptive activity \cite{holland_hidden_1996}.  In essence, the most important point of similarity to Baars’ theory is that incoming information is scanned for relevant items and those relevant items gain for a while the global resources of the cognitive system in order to produce the appropriate response.

The global workspace theory is a theory that aims to explain functional consciousness (in distinction from phenomenal consciousness). It starts from a more or less given cognitive system which presents certain complex behaviors and tries to explain how these behaviors are realized. Very briefly, the Global workspace model of consciousness operates as follows: many highly specialist and relatively simple cognitive functional modules are working in parallel, processing incoming information and competing on grabbing the central stage of the agent’s cognitive process. Once an item of information wins the competition it is globally broadcast to all modules, recruiting a great portion of computational resources of the agent to further attend to the relevant piece of information while other items are being suppressed. This grabbing of the central stage means the item was "brought to consciousness". But the glory of each such item is fleeting as importance decays in time and soon the whole sequence of competition, and global broadcast repeats itself.

Our starting point is fundamentally different. First, it is synthetic and not analytic i.e. it does not aim to explain an existing system (i.e. human cognition) but to construct a framework for an artificial cognitive process. There is no a-priori given shape to the system. In fact we aim for open ended emergence of complex cognitive functions. Second, as we aim to describe a scalable cognitive process, we need to address structures which are self similar at various scales. This is not a requirement of the global workspace model. Third, our framework aims to describe cognition as distributed within a diverse population of agents, while the original global workspace model, though utilizing massive parallelism at early stages, basically converges to a single stream of processing – the stream of functional consciousness. Fourth, a framework for an artificial cognitive development without positing a-priori shape of the system and its parts requires defining a process of boundary formation among different localities across scales.

We adapt from the global workspace model the basic idea as described above: in cognition, the selection for relevance (product of the attention mechanism) is the key for accessing the agent’s resources. Attention here is somewhat similar to functional consciousness in Baars’ and Franklin’s models in the sense that it indeed recruits the agent’s resources but it differs from it in some important points. The difference lies in the constitution of the selection mechanisms. The mechanisms we need for our framework must allow for a distributed operation across scales.
\subsection{Coalitions}
The simple competition and global broadcast model in Baars’ model is replaced by a more general concept of ad-hoc workspaces called coalitions. Coalitions are groups of interacting agents. A coalition implies the sharing and coordination of both information and semantic content among its participants that facilitates a collective cognitive function i.e. specialized selection for relevance (attention) and selection for action (intention) preformed collectively. Coalitions are consolidated by means of spreading activation and are constituted from the resources and know-how of the participating agents. Already Franklin \cite[pp. 14-15]{franklin_foundational_2006} mentions the forming of coalitions between competing modules in order to gain access to the global workspace. In our framework the idea of coalitions occupies a central role and is taken a few steps further:  1. In contrast to global attention in Baars’ theory, in our framework attention is distributed among many coalitions where each coalition is an ad hoc ‘workspace’ with specific capabilities. These ad-hoc workspaces are a product of self-organization within populations of interacting agents and are themselves components of higher level coalitions. 2. Interactions within coalitions involve not only exchange of information but also exchange of semantic content. 3. When a coalition self-organizes it actually performs as an ad-hoc cognitive agent with its own context and attention mechanism. This means that items of relevance and the cognitive agency that attends to them co-emerge and are co-dependent. This is a model more reminiscent of the way bacteria colonies coordinate feats of collective cognition \cite{ben-jacob_bacteria_2004,ben-jacob_self-engineering_2006}.  In our framework, the workspace from which actions ensue will always be multiple, distributed and spanning across a few scales. It will not be confined anymore to the constraints imposed by how biological brains evolved and developed in higher animals with central nervous system.

When a single cognitive agent is considered, cognition is basically local and amounts to a recurrent selection of relevant events and selection of a proper reaction to the selected events. With cognitive agents as components, what we aim at is twofold: (1) achieving an emergent distributed cognition within a diversified population of cognitive agents that self-organizes through their interactions and (2) achieving scalability, i.e. the self-organized coalitions and the processes involved in their emergence need to be such that they can recursively become the components of self-organization at progressively higher levels. Our definition of a cognitive agent in subsection \ref{sec:generic_cognitive_agent} serves accordingly two purposes: (1) it serves as a conceptual definition for a generic cognitive agent and (2) it describes the self-similar theme of our framework: at each level, the cognitive process is described with different kinds of events and actions but follows the same conceptual structure.

Here is how such distributed cognition system would be described given a population of interacting cognitive agents\footnote{The population need not be homogeneous. Agents may vary in their structure and function as long as they share the same platform of communicating events to each other.}:
\begin{enumerate}
	\item A subset of the population of agents identifies jointly a compound event as an item of relevance. A compound event is a set of contingent simple events that may or may not be causally connected or correlated. 
	\item The subset of agents forms a temporary coalition based on their shared "interest". Agents are reinforced to join coalitions they already joined in the past \cite{heylighen_self-organization_2013}. This is an abstracted version of the Hebbian rule \cite{hebb_0.1949_1968}: Agents that cooperated once will tend to cooperate again i.e. reinforce their connections. The recurrence of a coalition follows the recurrence of its activating compound event, when the coalition is reinforced and becomes more stable, it also binds, so to speak, the set of contingent events that stimulated its formation into a distinct gestalt. This is how compound events are consolidated into "compound items of relevance". The cognitive process therefore takes an active role in structuring the information flow from the environment. 
	\item The members of the coalition jointly select and coordinate their actions in response to the item of relevance that brought them together. While a coalition is active, the resources of the member agents are committed to the coalition and the connections among them stay stable.
	\item \label{itm:dismantled} Upon completion of the coordinated action the coalition is dismantled. The agents become again free to seek opportunistic coalitions but they ‘remember’ the coalitions they participated in and their strength of association to each, so they can join them again even without being specifically activated. According to this description, distributive cognitive activity is opportunistic as coalitions may form only once. But the reinforcement mechanisms mentioned in \ref{sec:generic_cognitive_agent} are necessary in order to ensure an ongoing tendency towards forming stable coalitions. Without a bias towards stability, the emergence of a hierarchy of more complex and capable super agents will be impossible. The tendency of an agent to participate in a coalition is dynamic. It starts with a slight bias in favour of joining any coalition. Once a specific coalition has been joined, the tendency to participate in that specific coalition is reinforced for all participating agents every time it is activated. However, all specific tendencies decay in time back to their initial strength. If a coalition is not activated frequently it is eventually forgotten.  The rationale behind this destabilization is double: first, a coalition that is not useful anymore because the circumstances of the environment have changed will slowly dissipate (be forgotten) releasing the constituent agents to form novel coalitions\footnote{It is assumed that every agent can effectively participate in a limited number of coalitions. If the number of coalition exceeds the limit, the rate of conflicts and malfunctions will increase and the whole process may collapse.}.  Second, the increase in scale necessarily implies an eventual decrease in the number of agents. If coalitions become too stable, the decrease in the number of agents necessarily implies a decrease in the variety of agents that can emerge at that scale. Decrease in variety will make the whole cognitive system less adaptable (with less options of configuration) at the higher levels, again because the principle of requisite variety. But if we take care to preserve plasticity at all scales by not letting coalitions to become unnecessarily stable, the variety at each level will not be bound anymore to the number of agents at that level as coalitions can reconfigure according to need\footnote{This ensures that there can be more kinds of agents than the number of actual agents at a given instance.}. The structure we propose therefore, never rigidify beyond a certain threshold and never loses adaptability as it becomes more structurally complex.
	\item Before being dismantled, the probability of an agent entering to a similar coalition in the future is updated according to the procedure outlined in item \ref{itm:dismantled}. Reinforcements can be further adjusted positively or negatively depending on the relative success or failure of the coordinated action selected by the coalition. Such adjustments may allow an adaptive or even evolutionary process to take place modifying the structure of the coalition but as a rule, unsuccessful coalitions will dissipate quicker while successful and influential\footnote{Influential means that the coalition is either highly connected or incorporated in many other coalitions. See following sections.} coalitions will increase in stability.
	\item Agents can participate in many coalitions, because usually their function in one coalition does not necessarily coincide temporally with their function in other coalitions. In cases that there is a conflict, an agent will be bound to the coalition it is most committed to (according to the level of consolidation of the said coalition as reflected by the connection strength of its participant agents and possibly additional parameters). Since coalitions will also have functional redundancy like the one characteristic in neural nets, they will be able to perform considerably well also in the absence of a few agents as discussed in \cite[pp. 60-62]{bechtel_connectionism_2002}.	
\end{enumerate}
\subsection{Coalitions as super agents}
A recurring coalition is in fact a cognitive agent too. The exact mechanisms of internal joining and coordinating are not specified at this level. As long as the behavior of the coalition can effectively be described in terms of selection for relevance and selection for effective action, this behavior complies to the definition of a cognitive process. When we want to emphasize the fact that an agent is constituted from a coalition of other agents we use the term super-agent. As super-agents emerge, the participating agents select together and respond to complex items of relevance in the environment. Moreover, because of their coordinated activities, their internal contexts become coupled and eventually form a joint context that can be said to belong to the super-agent\footnote{The coupling of contexts also means that each agent in a coalition becomes ‘aware’ to the inner state of other agents in the coalition as its selections and actions takes the inner contexts of other agents into account. In this sense, within a coalition there emerges a state of global awareness because of the said coupling.}. Super-agents can be said to discover patterns of events and form concepts of such compound events. The self-organization process that brings forth super-agents, also bestows structure on the environment, shaping domains within which these super-agents further interact (for a discussion of cognitive domains see \cite{von_glasersfeld_distinguishing_1997}). In the same manner emergent super-agents are capable of acting in a complex manner by producing mutually organized sequences of compound events. Super-agents as coalitions of many simpler agents are not entirely deterministic systems. Their characteristics may be given only in probabilistic terms and their inner structure i.e. the network of their constituent agents might itself be dynamic. In short, super-agents in this framework are characteristically plastic and complex.

It is also possible to describe a super-agent as a kind of episodic memory. Episodes are simply sequences of compound events that recur in a (more or less) specific order. The super-agent’s structure ‘remembers’ the compound events of relevance and associates each with a compound response. The complex sequences that constitute an episode trigger complex sequences of appropriate responses. This memory persists as long as it is reinforced by the recurrence of the relevant triggering events. Once the environmental circumstances cease to produce the episode, or, there is a change in the agent’s situation that renders the particular episode irrelevant, the coalition – the structure reinforced by this episode will start to dissipate and eventually disappear. The episode will be forgotten and the resources committed to respond to it will be released. Cognitive activity will proceed along alternative avenues.

To achieve scalability of the cognitive process it is required that the structure (and dynamics) of super-agents will facilitate their further incorporation into yet higher level coalitions. This requirement highlights the necessary structural self-similarity of the framework.
\subsection{Coalitions as a source of novelty}
A cognitive agent can always be described from two complementary points of view. The first point of view focuses on the structure and function of the cognitive agent in terms of input/output relations. From this point of view the agent is an independent dynamic system with more or less specific behavioral tendencies. Every input event, combined with the current state of the agent will make it follow a trajectory towards another state. Aspects of the trajectory and the new state will also determine possible output events. The second point of view focuses on the connections and coalitions a cognitive agent participates in. From this point of view the agent is described in terms of its capacities to affect and be affected by other agents in a heterogeneous population of agents \cite[pp. 1-6]{de_landa_philosophy_2011}. Agents can affect other agents via their actions and can be affected by other agents via their attention mechanism (which intentions of one agent gain the attention of another). A capacity to affect must involve another agent’s capacity to be affected. The important point here is that these capacities cannot be predicted from the dynamics of any single agent. Moreover, the variety within the population of agents gives rise to a combinatorial variety in capacities to affect and be affected which is translated to a variety of possible combined behaviors.

It follows that the behavior of coalitions of agents cannot be determined solely from the behaviors of the component agents. It might seem trivial at first sight but this unpredictability of capacities to affect and be affected is the source of the nearly inexhaustible innovation potential of coalitions. Any new capacity that emerges expands the horizon of additional new capacities. New capacities depend of course on the properties emergent in the dynamics of coalitions. When a coalition consolidates the various capacities that are expressed in the interactions among its participating agents give rise to the dynamism associated with the super-agent that emerges. The new agent’s properties are the resource of the capacities of the next level.

The combinatorial alignment of capacities among interacting agents is the process that critically produces novelty via the formation of new coalitions (super-agents) and stand therefore at the basis of the emergence of novel cognitive functions. From this perspective we can plausibly argue that our framework is capable, at least conceptually, to demonstrate open-ended innovation in cognitive capacities. 

\section{Extended framework of cognitive development}
	After putting together the conceptual ground we can now proceed to describe the future of the Web in terms of the cognitive development. We therefore first summarize main considerations of the above discussion as premises for the further development of the concept.

	\begin{description}
		\item  Agent -- We associate our concept of a generic cognitive agent with that of Burns and Engdahl \cite{burns_1._1998}. They consider what they call "agential capabilities" that in addition to what is implied by our framework also emphasizes the social construction of the identity (as embodied by its inner state) of every agent, meaning that the representation of self is always partially constructed from its social image. Any such socially constructed agent e.g. technological artifacts, social institutions and networks, worldviews, bacteria, anthills, human beings, etc. is considered therefore a cognitive agent. Cognitive agents are differentiated by the level of their cognitive development which also reflects their degree of individuation and sense-making capabilities\cite[p. 13]{burns_2._1998}.
		\item Sense-making -- We extend the concept of cognitive development from its human related context \cite{kegan_evolving_1982} to the generic cognitive agent. In this, more general framework, any process of sense-making extends the agent's cognitive competence and therefore is understood as cognitive development. By taking this approach we are not concerned about the question "whether an agent is cognitive or not?", but rather focus on questions such as "what is the current stage of an agent's cognitive development?",  "what is the agent's trajectory of development in terms of history and perceived future potential?". It is important to note that we use the concept of cognitive development not as a  notion of a sequence of these stages as it is used in the context of developmental psychology in humans but as a general model of intelligence expansion.
		\item  Sociotechnological system -- A cognitive agent at the largest scale and its future trajectories can be described in terms of cognitive development. 
	\end{description}
		
The application cognitive development to the sociotechnological system raises a few open questions which due to scope limits, we only touch here. In our model, a cognitive agent is recursively composed of smaller cognitive agents (e.g. cells $\rightarrow$ organs $\rightarrow$ bodies $\rightarrow$ groups $\rightarrow$ social institutions $\rightarrow$ societies $\rightarrow$ civilizations). It is reasonable to think that cognitive development at the highest scale (i.e. civilization) is influenced by the cognitive development processes at lower scales. The question arising here is what is the nature of this inter-scale interactions. A related question is to what extent research about eusocial insects informs our understanding, and perhaps more importantly, our projected direction regarding the future development of the Internet and Web especially when we think about the emergence of the Global Brain \cite[p. 47]{levy_collective_1997}, \cite{heylighen_return_2013}. 
		
\subsection{Society of Mind versus Mind of Society}

Minsky, explaining the concept of 'Society of Mind' states that 'words like \textit{living} and \textit{thinking} are useful for describing phenomena that result from certain combinations of relationship' \cite{minsky_society_1988}. When discussing collective consciousness (i.e. "Mind of Society"), Burns maintains that "collectives such as families, communities, administrative organizations, or states, are social agents and can be considered to possess agential capabilities" \cite{burns_1._1998}. It is obvious that these are different perspectives to the same phenomenon. When considering an agent, we are perceiving a boundary between individual and collective, between agent and environment, where the agent itself is already a collective.

\begin{figure}
	\centering
    	\includegraphics[scale=0.2]{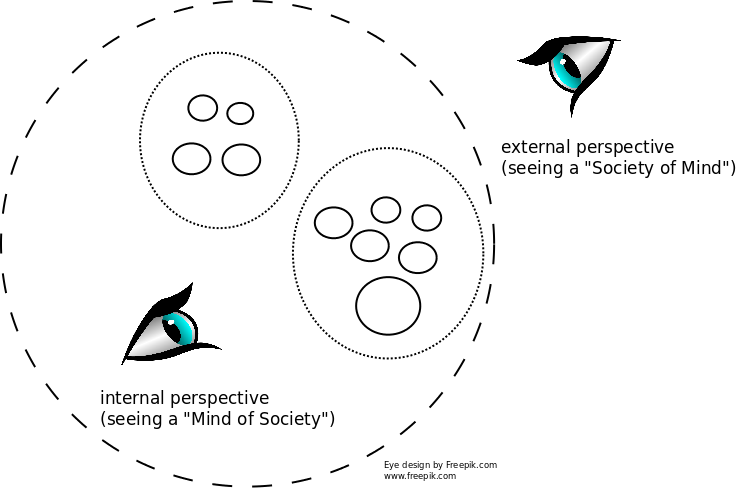}
  	\caption{Graphical illustration of "Mind of Society" vs. "Society of Mind" perspectives.}
  		\label{fig:mind_society_mind}
\end{figure}

When we analyze a cognitive agent as an individual and see a collective phenomena that makes this individual what it is, we take the perspective of 'Society of Mind' towards the boundary between individual and collective. But when we start from a collective phenomena and end up observing agential capabilities \cite{burns_1._1998}, we switch our perspective to 'Mind of Society' while still observing the same boundary (see Figure \ref{fig:mind_society_mind}).

On the ground of our framework we argue that social institutions e.g., social networks, government agencies or nation states are cognitive agents. We can observe boundaries of these entities both from the perspective of 'Mind of Society' (internal point of view) and 'Society of Mind' (external point of view). Furthermore we claim that worldviews are primary to social institutions and the reason of their existence. The cognitive development of these entities depends on their interactions e.g., in a form of networks of people or their digital identities. Therefore, the degree of cognitive development of the global sociotechnological system is determined by the dynamics of underlying ecosystem of worldviews.

\subsection{The formation of boundaries}

A boundary between an individual and a collective is that which separates self from other. It is important to remember that the boundary itself emerges in the course of cognitive development. Central to our approach is that the process of individuation comes prior to the individual as Kegan remarks: "Evolutionary activity involves the very creating of the object (a process of differentiation) as well as our relating to it (a process of integration)"\cite[p. 77]{kegan_evolving_1982}.

The "collectivity" of intelligence and individuation can be observed at two levels:
\begin{enumerate}
\item The individuation of each agent is driven by the surrounding environment which is a collective of agents. The process of the boundary formation between self and other does not depend solely on any of separated parts. Rather it is formed in the process of co-evolution and reflexive progressive determination when individual affects and is being affected by its environment.

\item The actions of a [pre-]individuated agent are driven by the competition and interaction among of internalized objects ("Society of Mind"). Within the framework of cognitive development of the Web we are concerned with the creative ecology of mental (as well as digital) representations and vicarious selectors, instead of physical embodiments.
\end{enumerate}

At every instance of the process, an agent \textit{holds} perceptional objects and \textit{is being held} as one of the perceptional objects of the surrounding agents. The interplay between these two aspects of "collectivity" form the individual-milieu dyad referred by Simondon \cite[p. 3]{simondon_genesis_1992}, which is equivalent to the boundary between scales of cognition. 

\subsection{Object relations theory}
The extended cognitive development process can be described as an ongoing balancing of \textbf{subject-object relationship} \cite{kegan_evolving_1982, pruyn_overview_2010} across the boundary of an agent. Subject-object relationship is the key mechanism describing how agents make sense of their sensory input. It is a recursive cycle of the following sub-processes (\cite{kegan_evolving_1982, pruyn_overview_2010, kegan_over_1995}):
\begin{itemize}
	\item \textit{Being subject to the sensory experience,} e.g. when a young child is sad, he/she does not relate to sadness as 'a temporary mental state', but \textit{IS sadness}.
	\item \textit{Separation of the sensory input/experience to an object i.e. creation of boundary with the environment.} Note that the theory does not assert that the subject differentiates \textit{a priori} between 'internal' and 'external' experiences. This differentiation is achieved during the cognitive development process with the goal to make sense of the totality of experiences. For example, a baby starting to recognize his/her mother as a separate individual.
	\item \textit{Holding an experience an object.} After separating an object and pushing it outside of its own representation of self, the agent remains with the \textit{internal mental representation} of that object.
	\item \textit{Using internal mental representation of objects for regulating experiences, including through actions}. Again, taking a child as an example, after internalizing the object of mother, a child can spend increasingly more time alone, finding comfort by relating to the internal rather 'the real' object.
\end{itemize}
It is our thesis is that analogous sub-processes can be adopted into our extended framework of cognitive development.

\begin{figure}[H]
	\centering
    	\includegraphics[scale=0.5]{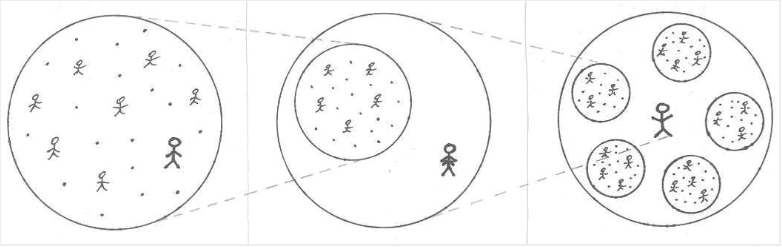}
  	\caption{A graphical illustration of cognitive development process as subject/object relationship, adapted from \cite{pruyn_overview_2010}}
  		\label{fig:cognitive_development}
\end{figure} 

\subsection{The "collectivity" of intelligence}

The mechanism of cognitive development applies both to what we call individual and collective intelligence, which we understand as different scales of the same phenomenon. The collective nature of the process can be approached from both perspectives of "Mind of Society" and "Society of Mind" (see Figure \ref{fig:mind_society_mind} on page \pageref{fig:mind_society_mind}) as well as collective individuation\cite{yuk_hui_collective_2013}.

Recursive levels of perceived boundaries between individual agencies and their collectives result in overlapping hierarchies of agents which emerge from interaction of simpler agents. This general structure is conceptualized by our framework for scalable cognition (see subsection \ref{sec:framework_of_scalable_cognition}). We posit scales of cognition not only because of their physical manifestations e.g. neurons  $\rightarrow$ brains $\rightarrow$ societies, or mental manifestations e.g. thoughts $\rightarrow$ memes $\rightarrow$ minds $\rightarrow$ cultures, but also because cognitive capabilities of agents at any boundary develop through \textit{interactions between scales}. Approached from the perspective of social construction of consciousness, "collective reflectivity emerges as a function of an organization or group producing and making use of collective representations of the self (we, our group, community, organization, nation) [..]"\cite{burns_1._1998,burns_2._1998}.

Taking a complementary perspective to intelligence as a 'Mind of Society'; merging the concepts of collective and individual intelligence; establishing its close ties with the environment and, most importantly, keeping in mind the definition of intelligence as a \textbf{process} of sense-making of the environment, we start to observe the relevance of the constructive-developmental approach to attempts to understand collective intelligence in the context of the development of the Web.

We understand cognition, intelligence (and actually, life) as a constant process of differentiation and integration, simultaneous growth and decay, separation from certain established structures and identification with others together with creation and destruction of internal representations of those structures. All these processes can be described under the wider concept of individuation \cite{simondon_genesis_1992,combes_gilbert_2013}. Individuation is defined in a much broader context than the psychological and encompasses physical systems, the evolution of organisms and life. By integrating the concepts of individuation and cognitive development we see them as describing a general principle of intelligence expansion across multiple scales of complex systems involving physical, biological, psychological as well as social and digital dimensions.

\section{The cognitive development of the Web}
We argue that the evolution of our the Web can be conceptualized as a cognitive development process. Furthermore, we propose that what we call a \textit{distributed governance in A World of Views}\cite{veitas_world_2013} is the next phase of the cognitive development of the human society whose current stage of cognitive development can roughly be compared to the one of an infant, meaning an intelligence dealing with reflexes, impulses and perceptions but having no needs and wishes distinguishable from these as yet. 

It is quite clear that the cognitive development of humanity through the World Wide Web will not follow the stages of human cognitive development which are derived from different collective patterns of sense-making that originate from human biological evolution and cultural history. According to our framework, all the scales of cognition are structurally coupled and co-evolve, therefore, the cognitive development of the global sociotechnological system cannot take place without a corresponding development of human agents and their social institutions. If human society is to reach its next phase of cognitive development, the lower scales of this cognitive agency i.e. human individuals and their social institutions, need to further evolve their sense-making capabilities as well. 

At the highest scale of cognitive agency that represents the individuation of human civilization as a whole, we put special emphasize on worldviews as the complex organizing patterns that characterize super-agents. We maintain that for the development of cognitive agents two operational contexts are necessary: the one represented by the "Mind of Society" perspective which highlights the motion towards integration and unity and the one represented by the "Society of Mind" perspective which highlights multiplicity as an ecology of challenge, co-evolution and differentiation that provides circumstances for the open-ended cognitive development of its participants. 

We envision the future sociotechnological system as a multiplicity of
unique, modular, open, co-evolving and competing worldviews. The development of such system will be largely influenced by current and future information
technologies of the Web and the Internet at large. In our vision, the Global Brain is more a platform for interaction of competing and cooperating multiple cognitive agencies than a singular unified entity that runs the risk of becoming a dead-end of intelligence expansion in the long run. We advocate therefore A World of Views and distributed governance as a facilitating platform of the Global Brain because they promise the open-ended nature and antifragility of the global sociotechnological system.

\subsection{Distributed governance}
Informed by the discussion above we argue that there is no single structure or model of the future sociotechnological system which can handle the increasing complexity and unpredictability of the world \cite{taleb_antifragile:_2012}. Therefore we see more importance in developing a global sociotechnological system towards a hypothesis testing engine not unlike the human cognitive system, but with much more permeable boundaries.

Hypothesis testing engine in the social realm means the ability to effectively respond to the needs of the global system for the new structures, institutions and initiatives in order to deal with the challenges at hand (e.g. climate change). Due to limited resources, massive hypothesis testing in the social domain requires an effective mechanism of both integrating and disintegrating social structures and institutions in a manner analogous to open source projects.
 
The implementation of these principles needs a change of basic conception of governance, namely, its rigid hierarchical nature. For an in-depth discussion of why hierarchical thinking is not suitable for near and far future see \cite{veitas_world_2013}.

We envision distributed governance as \textit{an ecology, medium and a mobilization system} for supporting the process of cognitive development of the global sociotechnological system. For that, distributed governance needs to facilitate four functions: 
\begin{enumerate}
	\item provide the necessary sustainable platform for a future society; 
	\item allow for the co-existence of diverse worldviews and agencies (humans, their collectives, social institutions and smart technological artifacts, including digital platforms);
	\item provide a medium for evolution of intelligence through communication, dialogue and co-evolution among diverse cognitive agents;
	\item establish an effective way of propagating successful experiments and containing failures which may be achieved by promoting a tight but dynamic coupling between different scales of the sociotechnological system (humans $\rightarrow$ worldviews $\rightarrow$ social institutions $\rightarrow$ sociotechnological system) enabling the on-going disintegration of obsolete social structures and the establishment of new ones in a distributed manner.
\end{enumerate}

In this essay we formulate broad principles of how we think our global sociotechnological system can and should be guided into the future. We do not provide specific recipes for regimes of governance because we think these are not practical. We emphasize instead the significance of the dynamics of change and we draw what we believe to be the correct direction of future trajectories.

We admit that much more work is needed for specifying the mechanisms through which distributed governance could be established taking the current status of the world as a point of departure. Nevertheless, formulating the conceptual principles of the desired system is the first necessary step.

\bibliographystyle{apalike}
\bibliography{bibliography_weaver,bibliography}

\begin{thebibliography}{}

\bibitem[Adams, 2010]{adams_why_2010}
Adams, F. (2010).
\newblock {Why we still need a mark of the cognitive}.
\newblock {\em Cognitive Systems Research}, 11(4):324--331.
\newblock Available from:
  \url{http://www.sciencedirect.com/science/article/pii/S1389041710000331}.

\bibitem[Ashby, 1958]{ashby_requisite_1958}
Ashby, W. (1958).
\newblock {Requisite variety and its implications for the control of complex
  systems}.
\newblock {\em Cybernetica}, 1(2):83--99.

\bibitem[Baars, 1993]{baars_cognitive_1993}
Baars, B. (1993).
\newblock {\em {A cognitive theory of consciousness}}.
\newblock Cambridge Univ Pr.

\bibitem[Baars, 2005]{baars_global_2005}
Baars, B. (2005).
\newblock {Global workspace theory of consciousness: toward a cognitive
  neuroscience of human experience}.
\newblock {\em Progress in brain research}, 150:45--53.

\bibitem[Bechtel and Abrahamsen, 2002]{bechtel_connectionism_2002}
Bechtel, W. and Abrahamsen, A. (2002).
\newblock {\em {Connectionism and the mind: Parallel processing, dynamics, and
  evolution in networks}}.
\newblock Wiley-Blackwell.

\bibitem[Ben-Jacob et~al., 2004]{ben-jacob_bacteria_2004}
Ben-Jacob, E., Aharonov, Y., and Shapira, Y. (2004).
\newblock {Bacteria harnessing complexity}.
\newblock {\em Biofilms}, 1(04):239--263.

\bibitem[Ben-Jacob and Levine, 2006]{ben-jacob_self-engineering_2006}
Ben-Jacob, E. and Levine, H. (2006).
\newblock {Self-engineering capabilities of bacteria}.
\newblock {\em Journal of the Royal Society Interface}, 3(6):197--214.

\bibitem[Burns and Engdahl, 1998a]{burns_1._1998}
Burns, T. and Engdahl, E. (1998a).
\newblock {1. The social construction of consciousness. Part 1: collective
  consciousness and its socio-cultural foundations}.
\newblock {\em Journal of Consciousness Studies}, 5(1):67--85.

\bibitem[Burns and Engdahl, 1998b]{burns_2._1998}
Burns, T.~R. and Engdahl, E. (1998b).
\newblock {2. The social construction of consciousness. Part 2: individual
  selves, self-awareness, and reflectivity}.
\newblock {\em Journal of Consciousness Studies}, 5(2):166--184.

\bibitem[Clark, 2012]{clark_whatever_2012}
Clark, A. (2012).
\newblock {Whatever next? Predictive brains, situated agents, and the future of
  cognitive science}.
\newblock {\em Behav. Brain Sci}.
\newblock Available from: \url{http://bi.snu.ac.kr/Courses/aplc12/3-2.pdf}.

\bibitem[Combes, 2013]{combes_gilbert_2013}
Combes, M. (2013).
\newblock {\em {Gilbert Simondon and the Philosophy of the Transindividual}}.
\newblock {MIT} Press.

\bibitem[{De Landa}, 2011]{de_landa_philosophy_2011}
{De Landa}, M. (2011).
\newblock {\em {Philosophy and simulation: the emergence of synthetic reason}}.
\newblock Continuum Intl Pub Group.

\bibitem[Franklin, 1995]{franklin_artificial_1995}
Franklin, S. (1995).
\newblock {\em {Artificial Minds}}.
\newblock {MIT} Press, Cambridge, Massachusetts.

\bibitem[Franklin, 2006]{franklin_foundational_2006}
Franklin, S. (2006).
\newblock {A foundational architecture for artificial general intelligence}.
\newblock In {\em {Advances in artificial general intelligence: Concepts,
  architectures and algorithms, proceedings of the {AGI} workshop 2006}}, pages
  36--57, Amsterdam. {IOS} Press.

\bibitem[Franklin, 2008]{franklin_ontology_2008}
Franklin, S. (2008).
\newblock {An Ontology For Cognition}.
\newblock Available from:
  \url{http://ccrg.cs.memphis.edu/tutorial/PDFs/AnOntologyforCognition.pdf}.

\bibitem[Franklin et~al., 1998]{franklin_ida:_1998}
Franklin, S., Kelemen, A., and McCauley, L. (1998).
\newblock {{IDA}: A cognitive agent architecture}.
\newblock In {\em {Systems, Man, and Cybernetics, 1998. 1998 {IEEE}
  International Conference on}}, volume~3, pages 2646--2651.

\bibitem[Hebb, 1968]{hebb_0.1949_1968}
Hebb, D. (1968).
\newblock {\em {0.(1949) The Organization of Behavior}}.
\newblock Wiley, New York.

\bibitem[Helbing, 2013]{helbing_globally_2013}
Helbing, D. (2013).
\newblock {Globally networked risks and how to respond}.
\newblock {\em Nature}, 497(7447):51--59.
\newblock Available from:
  \url{http://www.nature.com/nature/journal/v497/n7447/abs/nature12047.html}.

\bibitem[Heyes and Hull, 2001]{heyes_selection_2001}
Heyes, C.~M. and Hull, D.~L. (2001).
\newblock {\em {Selection Theory and Social Construct: The Evolutionary
  Naturalistic Epistemology of Donald T. Campbell}}.
\newblock {SUNY} Press.

\bibitem[Heylighen, 1999]{heylighen_growth_1999}
Heylighen, F. (1999).
\newblock {The growth of structural and functional complexity during
  evolution}.
\newblock {\em The evolution of complexity}, pages 17--44.

\bibitem[Heylighen, 2013a]{heylighen_return_2013}
Heylighen, F. (2013a).
\newblock {Return to Eden? Promises and Perils on the Road to a Global
  Superintelligence}.
\newblock In Goertzel, B. and Goertzel, T., editors, {\em {The End of the
  Beginning: Life, Society and Economy on the Brink of the Singularity}}.
\newblock Accepted for publication.

\bibitem[Heylighen, 2013b]{heylighen_self-organization_2013}
Heylighen, F. (2013b).
\newblock {Self-organization in Communicating Groups: The Emergence of
  Coordination, Shared References and Collective Intelligence}.
\newblock In Massip-Bonet, {\`A}. and Bastardas-Boada, A., editors, {\em
  {Complexity Perspectives on Language, Communication and Society}},
  {Understanding Complex Systems}, pages 117--149. Springer Berlin Heidelberg.
\newblock Available from:
  \url{http://link.springer.com/chapter/10.1007/978-3-642-32817-6_10}.

\bibitem[Holland, 1996]{holland_hidden_1996}
Holland, J. (1996).
\newblock {\em {Hidden order: How adaptation builds complexity}}.
\newblock Basic Books.

\bibitem[Holland, 2012]{holland_signals_2012}
Holland, J.~H. (2012).
\newblock {\em {Signals and Boundaries: Building Blocks for Complex Adaptive
  Systems}}.
\newblock The {MIT} Press.
\newblock Available from:
  \url{https://docs.google.com/uc?id=0B8Yh_B420F0sQmpDRlV3dVlxNG8&export=download}.

\bibitem[Kegan, 1982]{kegan_evolving_1982}
Kegan, R. (1982).
\newblock {\em {The evolving self: problem and process in human development}}.
\newblock Harvard University Press, Cambridge, Mass.

\bibitem[Kegan, 1995]{kegan_over_1995}
Kegan, R. (1995).
\newblock {\em {In Over our Heads - The Mental Demands of Modern Life}}.
\newblock Harvard University Press, Cambridge, Mass., {\'e}dition : reprint
  edition.

\bibitem[Klein et~al., 2006a]{klein_1._2006}
Klein, G., Moon, B., and Hoffman, R. (2006a).
\newblock {1. Making Sense of Sensemaking 1: Alternative Perspectives}.
\newblock {\em {IEEE} Intelligent Systems}, 21(4):70--73.

\bibitem[Klein et~al., 2006b]{klein_2._2006}
Klein, G., Moon, B., and Hoffman, R. (2006b).
\newblock {2. Making Sense of Sensemaking 2: A Macrocognitive Model}.
\newblock {\em {IEEE} Intelligent Systems}, 21(5):88--92.

\bibitem[L{\'e}vy, 1997]{levy_collective_1997}
L{\'e}vy, P. (1997).
\newblock {\em {Collective intelligence: mankind's emerging world in
  cyberspace}}.
\newblock Perseus Books, Cambridge, Mass.

\bibitem[Maturana, 1975]{maturana_organization_1975}
Maturana, H. (1975).
\newblock {The organization of the living: a theory of the living
  organization}.
\newblock {\em International Journal of Man-Machine Studies}, 7(3):313--332.

\bibitem[Minsky, 1988]{minsky_society_1988}
Minsky, M. (1988).
\newblock {\em {Society Of Mind}}.
\newblock Simon \& Schuster.

\bibitem[Pruyn, 2010]{pruyn_overview_2010}
Pruyn, P.~W. (2010).
\newblock {An Overview of Constructive Developmental Theory ({CDT})}.
\newblock Available from:
  \url{http://developmentalobserver.blog.com/2010/06/09/an-overview-of-constructive-developmental-theory-cdt/}.

\bibitem[Simondon, 1992]{simondon_genesis_1992}
Simondon, G. (1992).
\newblock {The Genesis of the Individual}.
\newblock In Crary, J. and Kwinter, S., editors, {\em {Incorporations}}, pages
  297--319. Zone, New York.

\bibitem[Systems,
  2013]{wefs_global_agenda_council_on_complex_systems_perspectives_2013}
Systems, W. G. A. C. o.~C. (2013).
\newblock {Perspectives on a Hyperconnected World - Insights from the Science
  of Complexity}.
\newblock Available from:
  \url{http://www3.weforum.org/docs/WEF_GAC_PerspectivesHyperconnectedWorld_ExecutiveSummary_2013.pdf}.

\bibitem[Taleb, 2012]{taleb_antifragile:_2012}
Taleb, N.~N. (2012).
\newblock {\em {Antifragile: Things That Gain from Disorder}}.
\newblock Random House.

\bibitem[Veitas and Weinbaum, 2013]{veitas_world_2013}
Veitas, V. and Weinbaum, D.~R. (2013).
\newblock {A World of Views}.
\newblock In Goertzel, B. and Goertzel, T., editors, {\em {The End of the
  Beginning: Life, Society and Economy on the Brink of the Singularity}}.
\newblock Accepted for publication.

\bibitem[von Glasersfeld, 1997]{von_glasersfeld_distinguishing_1997}
von Glasersfeld, E. (1997).
\newblock {Distinguishing the observer: An attempt at interpreting Maturana}.
\newblock {\em Towards an Ecology of Mind.[Originally published as: Die
  Unterscheidung des Beobachters: Versuch einer Auslegung. In: Riegas, V.,
  Vetter, C.(Eds.) Zur Biologie der Kognition. Suhrkamp, Frankfurt, 1990, pp.
  281-295]. http://www. oikos. org/vonobserv. htm}.

\bibitem[Wikipedia, 2014]{sensemaking_2014}
Wikipedia (2014).
\newblock {Sensemaking}.
\newblock {\em Wikipedia, the free encyclopedia}.
\newblock Available from:
  \url{http://en.wikipedia.org/w/index.php?title=Sensemaking&oldid=615797729}.

\bibitem[Willke, 2007]{willke_smart_2007}
Willke, H. (2007).
\newblock {\em {Smart Governance: Governing the Global Knowledge Society}}.
\newblock Campus Verlag.

\bibitem[{Yuk Hui} and {Harry Halpin}, 2013]{yuk_hui_collective_2013}
{Yuk Hui} and {Harry Halpin} (2013).
\newblock {Collective Individuation: The Future of the Social Web}.
\newblock {\em The Unlike Us Reader}, pages 103--116.
\newblock Available from:
  \url{http://digital-studies.org/wp/collective-individuation-the-future-of-the-social-web-by-yuk-hui-and-harry-halpin/}.

\end{thebibliography}
\end{document}